\documentclass{article}

\usepackage[preprint]{neurips_2025}

\usepackage[utf8]{inputenc} 
\usepackage[T1]{fontenc}    
\usepackage{hyperref}       
\usepackage{url}            
\usepackage{booktabs}       
\usepackage{amsfonts}       
\usepackage{pifont}        
\usepackage{nicefrac}       
\usepackage{microtype}      
\usepackage{xcolor}         
\usepackage{tabularx}
\usepackage{multirow}
\usepackage{amssymb}
\usepackage{graphicx}
\usepackage{geometry}
\usepackage{subcaption}
\usepackage{amsmath}
\usepackage{float}
\usepackage{marvosym}
\usepackage{verbatim}
\usepackage{longtable}
\usepackage{array}
\usepackage{xltabular}
\usepackage{ragged2e}
\usepackage{seqsplit}

\usepackage{tabularx}
\usepackage{array}
\usepackage{ragged2e}
\usepackage{booktabs} 

\usepackage{listings}
\usepackage{xcolor}

\definecolor{codegreen}{rgb}{0,0.6,0}
\definecolor{codegray}{rgb}{0.5,0.5,0.5}
\definecolor{codepurple}{rgb}{0.58,0,0.82}
\definecolor{backcolour}{rgb}{0.95,0.95,0.92}

\lstdefinestyle{mystyle}{
    backgroundcolor=\color{backcolour},   
    commentstyle=\color{codegreen},
    keywordstyle=\color{magenta},
    numberstyle=\tiny\color{codegray},
    stringstyle=\color{codepurple},
    basicstyle=\ttfamily\footnotesize,
    breakatwhitespace=false,         
    breaklines=true,                 
    captionpos=b,                    
    keepspaces=true,                 
    numbers=none,                    
    showspaces=false,                
    showstringspaces=false,
    showtabs=false,                  
    tabsize=2
}
\usepackage[T1]{fontenc}

\definecolor{deepblue}{rgb}{0,0,0.5}
\definecolor{deepred}{rgb}{0.6,0,0}
\definecolor{deepgreen}{rgb}{0,0.5,0}

\lstnewenvironment{python}[1][] 
{\lstset{language=Python, #1}} 
{}

\title{SCP: Accelerating Discovery with a Global Web of Autonomous Scientific Agents}
\author{Yankai Jiang\textsuperscript{*}\And
Wenjie Lou\textsuperscript{*}\And
Lilong Wang\textsuperscript{*}\and
Zhenyu Tang\and
Shiyang Feng\and
Jiaxuan Lu\and
Haoran Sun\and
Yaning Pan\and
Shuang Gu\and
Haoyang Su\and
Feng Liu\and
Wangxu Wei\and
Pan Tan\and
Dongzhan Zhou\and
Fenghua Ling\and
Cheng Tan\and
Bo Zhang\and
Xiaosong Wang\and
Lei Bai\and
Bowen Zhou \and \\
\textsuperscript{1}Shanghai Artificial Intelligence Laboratory \\
}
\date{November 2025}

\begin{document}

\maketitle


\begin{abstract}
Autonomous AI scientists are beginning to reason, experiment, and collaborate with human researchers by coordinating data, computational tools, AI agents, and laboratory instruments. Yet, most agentic science systems remain difficult to deploy beyond a single lab: they are bespoke, tightly coupled to specific workflows, and lack a shared \emph{protocol layer} that can unify heterogeneous dry–wet resources under a common, secure, and persistent scientific context.
We introduce \textbf{SCP: the Science Context Protocol}, an open-source standard designed to accelerate discovery by enabling a \emph{global web of autonomous scientific agents}. SCP is built on two foundational pillars:
\textbf{(1) Unified Resource Integration:} At its core, SCP provides a universal specification for describing and invoking scientific resources—spanning software tools, models, datasets, and physical instruments. This protocol-level standardization enables AI agents and applications to \emph{discover, call, and compose} capabilities seamlessly across disparate platforms and institutional boundaries.
\textbf{(2) Orchestrated Experiment Lifecycle Management:} SCP complements the protocol with a secure service architecture—comprising a centralized \textbf{SCP Hub} and federated \textbf{SCP Servers}. This architecture manages the complete experiment lifecycle (registration, planning, execution, monitoring, and archival), enforces fine-grained authentication and authorization, and orchestrates traceable, end-to-end workflows that bridge computational and physical laboratories.
Based on SCP, we have constructed a scientific discovery platform that offers researchers and agents a large-scale ecosystem of \textbf{1,600+} tool resources. Across diverse use cases, SCP transforms isolated agents and resources into interoperable building blocks. It facilitates secure, large-scale collaboration between heterogeneous AI systems and human researchers while significantly reducing integration overhead and enhancing reproducibility. By standardizing scientific context and tool orchestration at the protocol level, SCP establishes essential infrastructure for scalable, multi-institution, agent-driven science. The open-source SCP specification and reference implementation are available at \url{https://github.com/InternScience/scp}.
\end{abstract}

\section{Introduction}
AI scientists are emerging computational systems that can reason, experiment, and collaborate with human researchers throughout the scientific discovery process~\cite{lu2024ai,gao2025tooluniverse,wang2023scientific,tang2025ai,gao2024empowering,swanson2025virtual}. However, building such agentic systems in practice remains difficult. Most current deployments are bespoke, tightly coupled to a single laboratory or platform, and hard-wired to specific tools and workflows~\cite{swanson2024virtual,novikov2506alphaevolve,gao2025tooluniverse}. They typically expose ad-hoc interfaces to data repositories, simulation codes, and laboratory instruments, making it challenging to reuse components, reproduce workflows across institutions, or safely compose heterogeneous capabilities into end-to-end scientific pipelines in a common, secure environment. In particular, no widely adopted agentic system provides a unified protocol layer through which AI scientists, agents, and human-facing applications can interact with models, data, and instruments under a common, persistent scientific context~\cite{anthropic2024mcp}.

Recent advances in machine learning, large language models (LLMs) and laboratory automation have given rise to autonomous scientific agent platforms that can carry out research tasks with minimal human intervention~\cite{sourati2023accelerating,yang2024moose,weng2025deepscientist,gottweis2025towards,baek2025researchagent}.
Early demonstrations have largely centered on materials science and chemistry. For example, A-Lab~\cite{szymanski2023autonomous}, an autonomous materials synthesis laboratory, has demonstrated the power of AI-driven experimental autonomy by integrating computational databases, ML planning, and robotic execution. Over 17 days of continuous operation, it successfully synthesized 41 novel inorganic compounds out of 58 targeted ones, showcasing its high efficiency in materials discovery. Similarly, LLM-based agents like ChemCrow~\cite{m2024augmenting} (augmented with 18 expert-designed tools) and Coscientist~\cite{boiko2023autonomous} have autonomously planned and executed complex chemical syntheses, from organocatalysts to optimizing cross-coupling reactions. 

The paradigm is rapidly expanding to other critical scientific domains. In life sciences, platforms such as Origene~\cite{zhang2025origene} integrate structure prediction and sequence analysis tools to automate the design and engineering of functional proteins, accelerating the development of novel biocatalysts. For earth and environmental science, frameworks like EarthLink~\cite{guo2025earthlink} connect heterogeneous data sources and knowledge models to enable autonomous reasoning for tasks ranging from climate analysis to ecological impact assessment. At a systemic level, multi-agent architectures like InternAgent~\cite{team2025internagent} coordinate specialized sub-agents to manage end-to-end research tasks—from literature review to experimental validation—demonstrating the potential for orchestrating complex, cross-disciplinary workflows.
Concurrently, broader frameworks are being developed to integrate the entire research lifecycle. Agent Laboratory~\cite{schmidgall2025agent} is a system of specialized LLM agents that autonomously generates comprehensive research outputs from a human-provided idea. Kosmos~\cite{mitchener2025kosmos} employs a structured world model for parallel multi-agent reasoning over data and literature. Collectively, these efforts underscore a clear trajectory: when equipped with domain-specific tools and structured protocols, AI agents can automate critical research steps and execute end-to-end investigations at a scale and speed beyond the reach of individual human researchers.

The foregoing examples, drawn from diverse fields, illustrate \textbf{the emergence of an interconnected “web” of autonomous scientific agents}, each capable of carrying forward portions of the scientific method. However, existing orchestration frameworks~\cite{openaiIntroducingDeepResearch2025,futurehousePlatform2025,deepseBohrium2025,opensciencelabSCP2025} and tool registries~\cite{gao2025tooluniverse,ding2025scitoolagent} operate at the level of individual applications and do not provide a protocol-level abstraction for scientific context and lifecycle management. Moreover, true cross-system and cross-institution collaboration among scientific agents remains a challenge. Key obstacles include establishing a consistent shared context across platforms, managing the state of ongoing experiments, and enforcing data access and permission boundaries. Existing standards (\textit{e}.\textit{g}. the Model Context Protocol) do not fully resolve these issues, making it difficult for heterogeneous agent systems to interoperate seamlessly. 

\begin{figure}[t]
    \centering 
    \includegraphics[width=\textwidth]{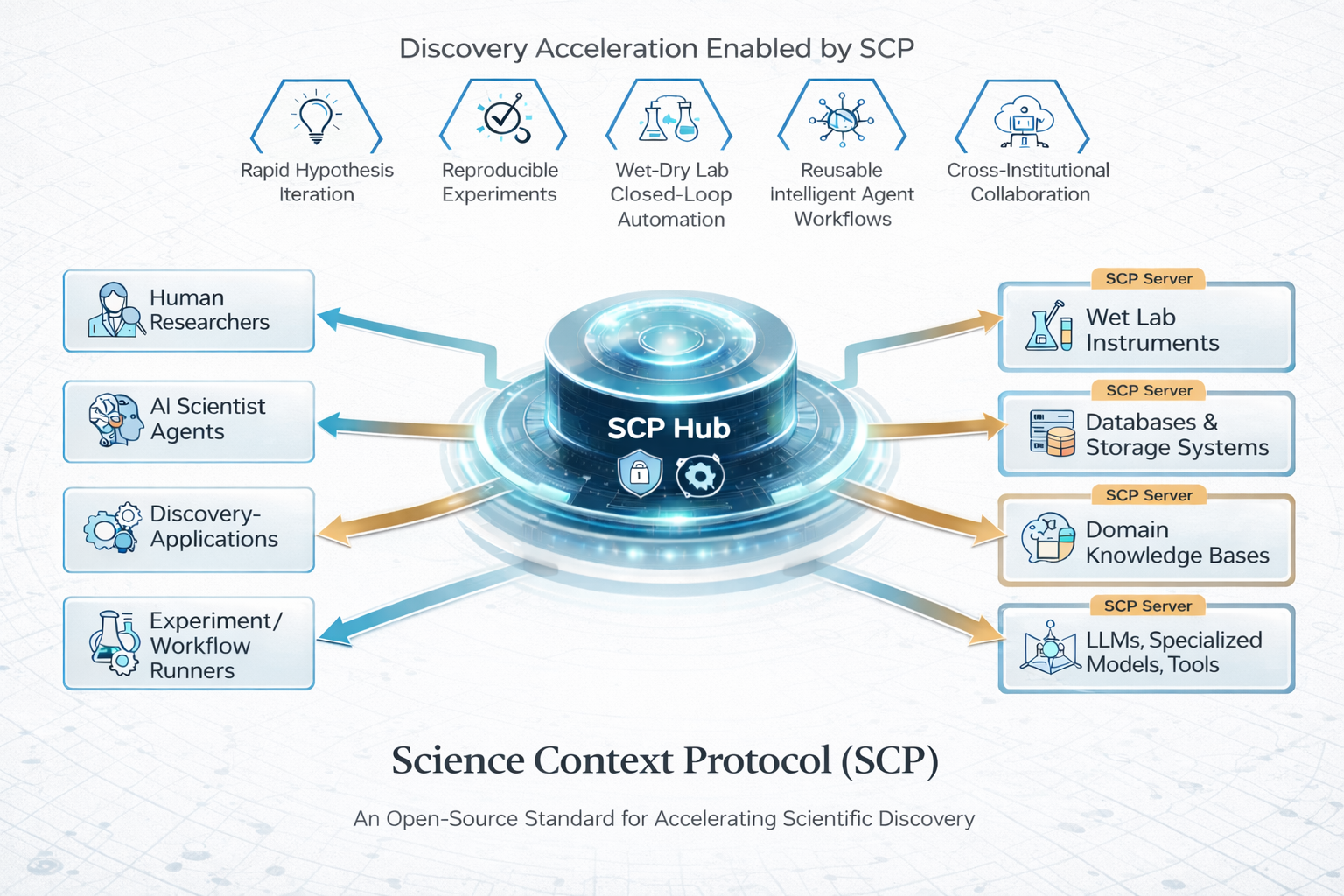} 
    \caption{SCP overview. The Science Context Protocol (SCP) is an open-source standard specifically designed to accelerate scientific discovery. By establishing a standardized connectivity framework, it enables efficient interaction between discovery-oriented applications and external research assets—such as laboratory instruments, databases, knowledge repositories, large language models (LLMs), specialized computational models, tools, and APIs. SCP aims to foster a hybrid dry-wet, multi-institution collaborative research paradigm and serve as a novel support platform to enable the collaborative evolution of researchers, research tools, and research subjects in a new era of multi-agent-driven scientific investigation and discovery.} 
    \label{fig:intro} 
\end{figure}

To address this infrastructure gap, we propose the \emph{Science Context Protocol (SCP)}, a unified framework that enables secure, context-aware collaboration among researchers, tools, and autonomous scientific agents. By linking distributed AI agents into a cooperative global network, SCP aims to accelerate discovery through collective intelligence and resource sharing, laying the foundational infrastructure for a collaborative, accelerated, and globally connected paradigm of scientific discovery.
SCP builds on model--tool interaction protocols and tool--centric ecosystems, and further extends them in three key directions that are critical for scientific use. First, SCP introduces a centralized \textbf{SCP Hub} that maintains global scientific context, enforces per-experiment authentication and authorization policies, and orchestrates calls to a federated set of \textbf{SCP Servers} that expose tools, data resources, and physical instruments. 
Beyond simple request routing, the  Hub embeds an intelligent orchestration system that plans and supervises multi-stage workflows spanning computational tools, domain-specific agents, and both dry and wet laboratory operations. It continuously senses the current environment to detect available resources (tools, datasets, and agents), and, given a high-level experimental goal, automatically synthesizes candidate tasks. For each experiment, the Hub ranks the top-$k$ executable plans and surfaces them to the user or AI scientist together with decision rationales such as dependency structure, expected latency, experimental risk, and cost estimates. The orchestration layer is coupled to an internal AI governance module that performs conflict detection and resource forecasting, issuing early warnings when workflows over-constrain instruments, data, or compute budgets. All of these capabilities are exposed through a programmatic API, allowing external AI scientists and applications to request end-to-end plans rather than manually scripting low-level tool calls. The Hub thus acts as the ``brain'' of the system: it parses high-level intents from AI scientists or human users, decomposes them into multi-step experimental plans, and coordinates execution across dry and wet resources while maintaining a persistent audit trail. Second, SCP generalizes the notion of a ``tool'' to include not only software functions and models but also laboratory devices, workflow engines, and composite multi-agent procedures. Each SCP Server uses a common specification schema to describe the capabilities it offers, their parameters, side effects, and security requirements, allowing clients to reason about and safely compose heterogeneous resources. Third, SCP explicitly models the lifecycle of scientific tasks, from experiment registration and planning through execution, monitoring, and archival, providing first-class abstractions for experiment identifiers, context objects, logs, and provenance.
Complementing the orchestration layer, SCP Hubs also drive an automated execution subsystem that instantiates chosen plans, performs secondary validation of preconditions, and manages task state over time (including running, pausing, resuming, and terminating experiments). This subsystem provides asynchronous anomaly notifications and live monitoring views that track tool status, data flows, and resource utilization, and it can trigger predefined fallback strategies when failures or abnormal patterns are detected.

SCP is implemented as an open-source reference platform that supports both local and remote deployment. SCP Hubs and Servers communicate over secure channels and can be integrated with existing identity and access management systems to respect institutional boundaries. This architecture enables AI scientists and scientific applications to orchestrate multi-institution workflows: for example, a single SCP experiment may combine literature retrieval, simulation on a remote high-performance computing cluster, and closed-loop control of robotic laboratory instruments, all expressed within one coherent context and governed by a common policy. Early deployments of SCP in scientific discovery platforms demonstrate that this protocol can break down data and capability silos, turn curated scientific corpora into ``data as a service'' for AI models, and support safe collaboration between heterogeneous agents and human researchers.

\section{SCP Architecture}
\label{sec:architecture}

\subsection{Core Components}
\label{subsec:core-components}

SCP adopts a hub--and--spoke architecture in which a centralized \emph{SCP Hub} coordinates a federation of distributed \emph{SCP Servers} (edge servers), user-facing \emph{SCP Clients}, and a set of auxiliary messaging and storage components. Together, these elements form a uniform connectivity fabric that links AI scientists and scientific applications to heterogeneous resources across laboratories and institutions. Conceptually, this design extends model--tool protocols such as MCP~\cite{anthropic2024mcp} from generic tool invocation to the full lifecycle of scientific experiments, including experiment registration, planning, execution, and provenance tracking.

On top of the basic MCP-style client--server interaction model~\cite{anthropic2024mcp}, SCP introduces four major extensions tailored to scientific workflows.  
(i) \textbf{Richer experiment metadata.} SCP defines a first-class experiment context that records a persistent experiment identifier, experiment type (dry, wet, or hybrid), high-level goals, data storage URIs, and configuration parameters. This structured context supports end-to-end traceability, versioning of experimental runs, and integration with institutional data-governance policies.  
(ii) \textbf{Centralized SCP Hub.} Unlike purely peer-to-peer protocols, SCP designates the Hub as a global registry for all experiment-facing services (data sources, computational models, laboratory instruments, and composite agents). The Hub manages service discovery, lifecycle management, experiment memory, and OAuth2.1-based authentication and authorization for experiments and users.  
(iii) \textbf{Intelligent workflow orchestration.} Beyond single tool calls, SCP adds an \emph{experiment-flow API} layered over conventional agent APIs~\cite{gao2025tooluniverse}. This API allows the Hub to synthesize and recommend candidate multi-step workflows---including resource allocation and follow-up actions---from a high-level experimental goal, enabling AI-driven generation and execution of complete experimental protocols.  
(iv) \textbf{Wet-lab device integration.} SCP extends the notion of a tool to cover real laboratory equipment by standardizing device drivers and capability descriptions. This enables dry (computational) and wet (physical) experiments to be composed into unified workflows under one protocol, making robotic platforms, analytical instruments, and in-silico models equally addressable through SCP.

We now describe the responsibilities of the core components in this architecture.

\begin{figure}[t]
    \centering 
    \includegraphics[width=\textwidth]{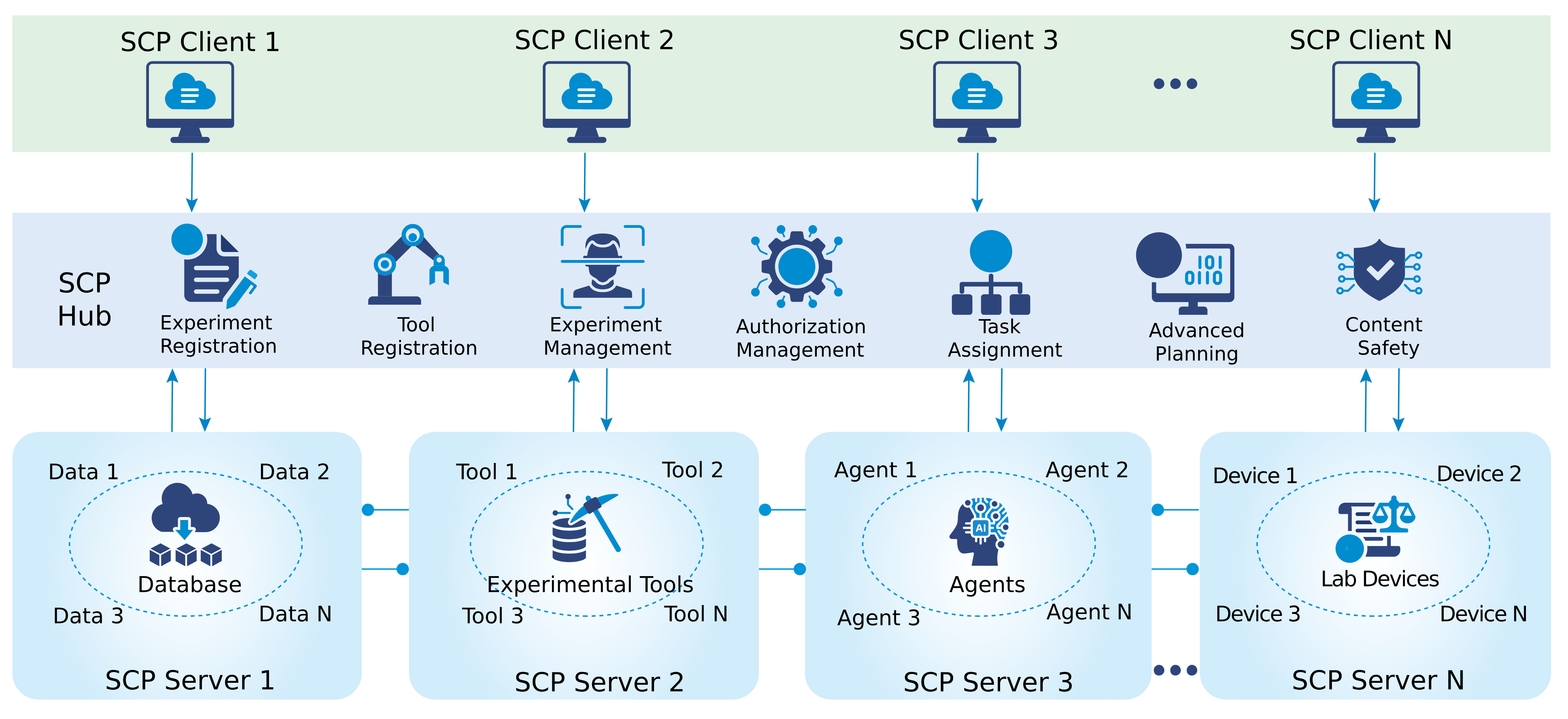} 
    \caption{SCP architecture overview. The SCP Hub coordinates interactions between user-facing clients (top) and various SCP \emph{edge servers} (bottom) that interface with laboratory instruments, databases, AI models, and other tools. Researchers interact with the system through an SCP client application, which communicates with the Hub. The Hub manages experiment context, planning, and task scheduling across the network of tools. Each SCP edge server registers available devices or services with the Hub, executes tasks on those resources, and streams results back to the Hub in real time. This design enables a seamless flow of information and commands between human researchers, AI-driven agents, and physical lab equipment under a unified protocol.} 
    \label{fig:architecture} 
\end{figure}

\subsubsection{SCP Hub (central orchestrator and protocol authority)}

The SCP Hub is the ``brain'' of the system. It maintains the global registry of tools, datasets, agents, and instruments; handles service discovery; performs task dispatch and tracking; and implements protocol-level security and governance. Given a high-level request from an AI scientist or user-facing application, the Hub interprets the intent under the current experiment context, decomposes it into a sequence of concrete experimental tasks, and coordinates their execution across multiple Servers.

\paragraph{Intelligent orchestration of domain knowledge, tools, agents, and experiments.}
As outlined in the Introduction, a key innovation of SCP is the intelligent orchestration layer embedded in the Hub. When a user submits a complex scientific task, the Hub first uses AI-driven intent-analysis models to translate natural-language instructions into a set of candidate task graphs involving tools, datasets, agents, and both dry and wet experimental operations. It then evaluates these candidates against the current environment: available tools and instruments, dataset readiness, user permissions, and resource budgets. For each experiment, the Hub ranks the top-$k$ executable plans and surfaces them---together with dependency structure and coarse estimates of latency, cost, and experimental risk---to the human researcher or higher-level AI scientist for selection. Alternatively, users who prefer direct control can bypass the high-level intent analysis and manually compose their workflows using the underlying APIs or graphical interfaces. The intelligent orchestration mechanism is exposed via the experiment-flow API, which is layered on top of traditional agent APIs~\cite{gao2025tooluniverse,anthropic2024mcp} and enables AI scientists to request end-to-end workflows rather than manually chaining individual tool calls.

\paragraph{Fine-grained repeatable protocol-level workflow specification.}
Once a candidate plan has been selected, the Hub compiles it into a fine-grained, protocol-level workflow specification, represented as a structured JSON task graph. Each node in this graph corresponds to a single operation---such as calling a data-cleaning pipeline, running a simulation, executing a trained model, or actuating a robotic liquid handler---and is annotated with the tool identifier, input parameters, expected outputs, and explicit dependencies on previous steps. This JSON representation standardizes the experimental protocol at the level of the SCP architecture: it serves as the contract between AI scientists, Servers, and instruments, and as the canonical record of ``what was planned'' for an experiment. The Hub versions and stores these workflow specifications as part of the experiment memory, enabling replay, comparison across runs, and downstream auditing of experimental decisions.

\paragraph{Automated execution, feedback, and interaction.}
Given a validated workflow specification, the Hub enters an automated execution phase. It dispatches each step in dependency order to the appropriate SCP Server, supplying the required parameters and experiment-context metadata. Servers execute the requested operations and stream back status updates and results. The Hub tracks task states (running, paused, completed, failed) over time and logs server response latency for each step, providing live monitoring views of tool status, data flows, resource utilization, and performance metrics.

Critically, the Hub implements a built-in validation and feedback loop. For data-centric steps, it can verify that returned values conform to expected schemas or basic sanity checks; for physical operations, it inspects device acknowledgements and sensor readings to confirm that actions were successfully applied. If a step fails validation or an anomaly is detected (for example, repeated tool errors, inconsistent outputs, or unexpected resource consumption), the Hub triggers an exception-handling policy: issuing warnings, pausing or aborting downstream steps, and optionally rolling back prior actions when safe to do so. These rollback and recovery policies are particularly important in multi-step, dry–wet workflows, where early errors can otherwise propagate and invalidate subsequent measurements. Through this dispatch–monitor–validate–rollback loop, the Hub acts as a robust execution coordinator that keeps large, multi-agent scientific workflows on track.

\subsubsection{SCP Servers (edge nodes)}

An SCP Server connects to and manages local scientific resources within a given environment (\textit{e}.\textit{g}., a laboratory, data center, or cloud account). Typical resources include experimental instruments, domain-specific models, databases, data-processing pipelines, and higher-level agents. The Server is responsible for registering these resources as SCP tools, exposing their capabilities and parameter schemas through a common specification format, and enforcing local access-control policies.

At runtime, each SCP Server receives task directives from the Hub and invokes the appropriate tools or devices to execute specific experimental steps. It streams intermediate states and final results back as structured messages, enabling the Hub to maintain a global view of progress. On startup, a Server automatically registers itself and its exported tools with one or more Hubs, making its local capabilities discoverable to AI scientists and clients across the SCP ecosystem. To facilitate integration, the SCP Hub provides a set of registration interfaces through which users can submit their own Servers—for example, via the Intern-Discovery platform~\cite{intern-discover-platform}—to be incorporated into the unified management of the platform’s SCP Hub. Servers also perform continuous health monitoring, periodically reporting device status, model readiness, and resource utilization to the Hub. These reports inform global scheduling and failover decisions made at the Hub level.

\subsubsection{SCP Clients}

SCP Clients provide the interface layer for human researchers, AI scientists, and higher-level applications. A Client authenticates users against the Hub, retrieves the list of tools, agents, datasets, and instruments that are visible under the user’s permissions and project context, and offers interaction surfaces for experiment design and execution. In simple cases, a Client can present a catalog-style interface for discovery and direct invocation of individual tools. In more advanced scenarios, it exposes graphical or programmatic builders for composing high-level experimental goals, which are then handed off to the Hub’s orchestration layer.

Every invocation issued by a Client carries the user’s credentials and the relevant experiment identifier, allowing the Hub to enforce fine-grained access-control policies and attribute actions to specific users or AI agents. This design ensures that, even as SCP connects a growing ecosystem of Servers across institutional boundaries, experiment-level security and provenance remain centralized and auditable at the Hub.

\subsection{High-Level Usage Patterns and Workflows}\label{sec:usagepatterns}
In this section, we outline how scientists, developers, and infrastructure architects could interact with SCP to carry out and support complex experiments. At a high level, SCP provides a unifying interface for researchers to design and execute experiments, for developers to integrate new tools and instruments, and for architects to deploy a scalable, secure infrastructure. Figure~\ref{fig:architecture} illustrates the overall SCP platform architecture, showing how users, intelligent agents, and laboratory devices are connected via the centralized SCP Hub. Based on the above architecture, the user-facing process of conducting an experiment with SCP proceeds through several stages:

\textbf{(1) Experiment Query and Design (Researcher Interaction):} A scientist begins by formulating an experimental request or query using an SCP client interface. This could take the form of a structured query (selecting from available protocols) or a natural-language request describing the experiment’s goal. The SCP client helps the researcher by showing the inventory of available tools and data resources (filtered by the user’s permissions) and providing a user-friendly interface for composing an experiment workflow. For example, a biologist might ask the system to “find the optimal protocol to synthesize compound X and measure its activity,” or manually select a series of laboratory actions to perform. The request is sent to the SCP Hub along with relevant metadata (experiment name, type, priority, etc.), establishing a new experiment context within the platform.

\textbf{(2) Protocol Generation and Planning (AI Agent \& Hub Orchestration):} Upon receiving the query, the SCP Hub interprets the researcher’s intent and generates a detailed experimental protocol. This process can involve intelligent planning agents (e.g. an LLM-based planner) that analyze the request and propose a sequence of steps to achieve the goal. The SCP Hub’s intent recognition module parses the input to identify high-level tasks and constraints. It then evaluates available methods and resources registered in the system, possibly consulting domain-specific models or knowledge bases for recommendations. The outcome is one or more candidate protocols expressed in a structured JSON-based format that enumerates the required steps, tools, and parameters for the experiment. Notably, the Hub may suggest multiple ways to proceed — for instance, different instruments or methods to accomplish a step — and present the top three execution plans to the user for review. Each proposed plan includes metadata (estimated duration, resource usage, predicted outcomes) to help the researcher make an informed selection.

\textbf{(3) Selection and Orchestration of Domain Knowledge and Tools (Researcher \& Hub):} The researcher reviews the suggested protocols and selects a plan (or manually edits it) to execute. Once a plan is confirmed, the SCP Hub orchestrates its execution. It decomposes the high-level protocol into a series of discrete actions and assigns these tasks to the appropriate SCP edge servers that manage the required tools or lab devices. For example, if the protocol calls for a temperature-controlled reaction followed by spectral analysis, the Hub will route the heating step to a connected thermostatic reactor device and the analysis step to a spectroscopy tool, via their respective SCP servers. All task assignments are done through the standard SCP interface, encapsulating each command and its parameters as defined by the protocol specification. The Hub coordinates scheduling, taking into account the availability and status of each device, and initiates each step in turn or in parallel as needed. Crucially, the researcher can monitor this orchestration in real time: the SCP client provides live updates or a dashboard view of which tasks are running, completed, or pending. The unified protocol ensures that whether a step is executed by a robotic lab instrument, an AI model, or a data processing pipeline, it is invoked in a consistent manner and tracked under the same experiment context.

\textbf{(4) Execution on Laboratory Devices and Tools:} Each SCP edge server receives tasks from the Hub and carries out the specified action on the local resource under its control. The SCP server software at the lab side translates the high-level protocol command into the device-specific operation (for instance, moving a robotic arm, running a simulation, calling an external API, or querying a database). Developers have extended the SCP servers with a library of device drivers and tool adapters, so that most actions simply call pre-defined routines exposed by the device’s API. During execution, the edge server streams the action’s progress and any intermediate data back to the Hub. This could include logging information, sensor readings, or partial results (e.g., an image just captured by a microscope). The Hub aggregates these updates and makes them available to the researcher in real time, ensuring transparency and allowing human oversight if needed. 

\textbf{(5) Feedback, Analysis, and Iteration:} As the experiment progresses, the SCP Hub closes the loop by analyzing results and feeding them back into the context of the running experiment. If an intelligent agent is steering the experiment, it can use the incoming data to decide on-the-fly adjustments. For example, an AI planning agent might detect that a reaction’s yield is below the expected threshold and suggest modifying the temperature or trying an alternative catalyst in a subsequent run. The researcher can either approve these adaptive changes or intervene with new instructions if necessary. Once all steps (including any conditional or looped steps) are completed, the SCP client presents the final results to the scientist. The outcomes might include processed data, graphs, lab notes, or even a draft report generated by an AI assistant. Because SCP assigns uniform identifiers to every data object and records every action in the protocol, the entire experiment is traceable and reproducible. Researchers (or other team members) can review the JSON protocol log to understand exactly what was done and to repeat the experiment in the future, satisfying a key requirement for scientific rigor.

The above workflow is made possible by developers who extend the SCP platform with new tools and capabilities. From a developer’s perspective, using SCP involves writing adapters or drivers that wrap instruments and software tools behind the SCP protocol. For example, a developer can implement a new device action by subclassing the SCP server’s device interface or using a decorator to register a function as a tool action. Once the code for a new tool is written, the developer deploys it on an SCP edge server. The edge server automatically registers its tools and services with the central Hub, advertising what actions it can perform. This modular design means developers can continually add support for new hardware or analysis routines without modifying the core of SCP. Scientists immediately gain access to those new capabilities via the unified interface. In practice, developers also use SCP’s client APIs to script complex tasks or to build higher-level applications. They might write Python scripts that connect to the SCP Hub to programmatically submit experiment protocols or retrieve results, which is useful for integrating SCP into custom pipelines or GUIs. The SCP project’s open-source repository provides documentation and examples for developers, including how to define custom actions and ensure they conform to the protocol’s JSON schema.

Ultimately, the coordination provided by SCP – from low-level device control up to high-level experiment planning – allows researchers to focus on scientific questions while the software handles the complex choreography between AI agents and laboratory infrastructure. This synergy between scientists, AI agents, and automated lab systems illustrates a new paradigm of conducting research, one that is markedly more efficient, collaborative, and adaptive than traditional methods~\cite{BioMARS2025,omni2025}.

\section{Tool Assets: The SCP-Based Intern-Discovery Platform}
Building on SCP, we have constructed a large-scale, highly diverse tool assets on Intern-Discovery platform~\cite{intern-discover-platform}. This platform integrates over 1,600 interoperable tools spanning databases, computational utilities, and specialized models across multiple scientific domains, forming one of the most comprehensive tool ecosystems reported for scientific agent systems. The complete scp server list and tool list is provided in Table~\ref{tab:scpserver} and Table~\ref{tab:toollist} respectively in the Appendix.

From a functional perspective, the platform covers the full spectrum of scientific workflows, including biological and chemical database retrieval (e.g., UniProt, InterPro, PDB, NCBI), structure and sequence processing, molecular modeling, data preprocessing, numerical computation, and machine learning inference. From a disciplinary standpoint, these tools collectively support a broad range of fields such as molecular biology, protein engineering, bioinformatics, chemistry, materials science, and medical research.

Unlike prior agentic or tool-augmented frameworks that typically rely on a limited and task-specific toolset, the SCP protocol provides a unified, standardized, and extensible interface that enables large language model agents to seamlessly access and orchestrate this massive collection of heterogeneous tools through a common JSON-based schema and a centralized hub. This large-scale, protocol-enabled integration not only significantly expands the action space and problem-solving capacity of scientific agents but also facilitates cross-domain reasoning and complex multi-step workflows that are difficult to achieve with conventional systems. As a result, the SCP-based platform establishes a robust and general-purpose foundation for scalable, tool-driven scientific intelligence, demonstrating clear advantages in flexibility, coverage, and long-term extensibility.

\begin{figure}[htbp]
    \centering
    
    \begin{subfigure}[b]{0.48\textwidth}
        \centering
        \includegraphics[width=\textwidth]{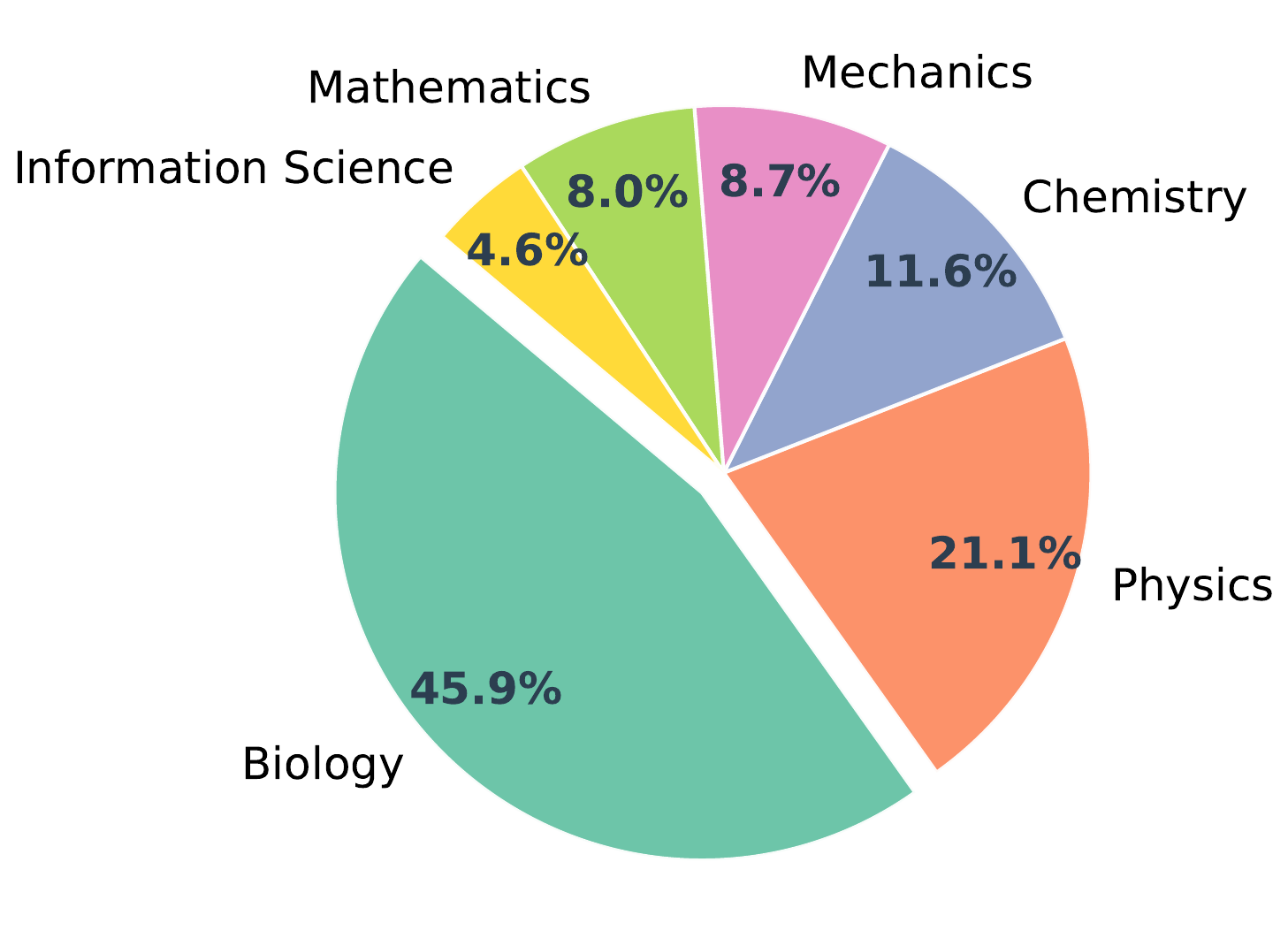}
        \caption{Disciplinary Distribution of Tools on the Intern-Discovery platform~\cite{intern-discover-platform}.}
        \label{fig:disciplinary}
    \end{subfigure}
    \hfill
    \begin{subfigure}[b]{0.48\textwidth}
        \centering
        \includegraphics[width=\textwidth]{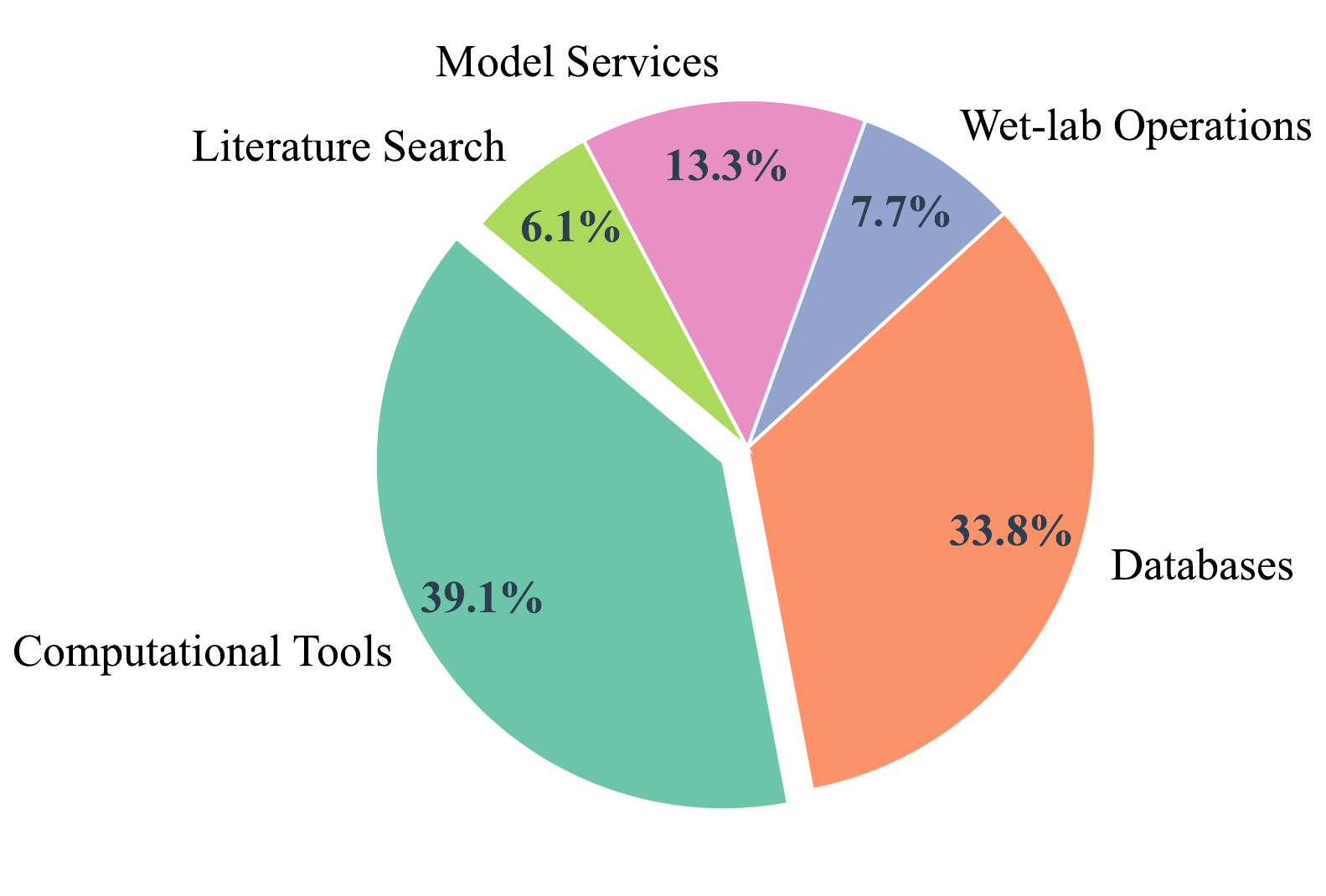}
        \caption{Functional Category Distribution of Tools on the Intern-Discovery platform~\cite{intern-discover-platform}.}
        \label{fig:functional}
    \end{subfigure}
    \caption{Distribution of disciplines and functions of tools available on the Intern-Discovery platform~\cite{intern-discover-platform}. The tool collection is continuously updated and expanded.}
    \label{fig:distribution}
    \small
\end{figure}

In Figure~\ref{fig:disciplinary} and Figure~\ref{fig:functional}, we present the disciplinary distribution and functional categorization of the tools available on the platform. The tools encompass a broad spectrum of disciplinary categories.

\begin{itemize}
    \item \textbf{Biology and Related Technologies (45.9\%):} This category covers the complete biomedical research pipeline, ranging from genomics and proteomics to drug discovery and disease research.
    \item \textbf{Physics (21.1\%):} This includes major branches of physics, encompassing optics and electromagnetism, thermofluids, electromagnetics, and fundamentals across various sub-disciplines.
    \item \textbf{Chemistry (11.6\%):} This comprises chemical molecular databases, chemical and reaction computations, and multiple chemistry branches such as computational chemistry and physical chemistry.
    \item \textbf{Mechanics and Materials Science (8.7\%).}
    \item \textbf{Mathematics (8.0\%).}
    \item \textbf{Information Science and Computing Technology (4.6\%).}
\end{itemize}

This distribution closely reflects the trend in modern scientific research, particularly the deep integration of experimental disciplines (\textit{e}.\textit{g}., biology, materials science) with computational methodologies.

\section{Case Study} 
\subsection{Case Study 1: Automated Experimental Protocol Design and Execution} A researcher begins by submitting a high-level experimental objective (for example, ``design and run a PCR protocol to verify a gene knockout'') to the SCP hub. The SCP hub orchestrates this request by distributing tasks across its network of service nodes. A planning agent translates the objective into the standardized SCP JSON format, and the Thoth Server is invoked as a specialized SCP service node to generate the detailed laboratory protocol \cite{sun2025unleashing}. Thoth processes this input and returns a structured JSON protocol that describes each step of the experiment, including reagents, quantities, and timing. Next, the SCP hub proceeds with execution. It forwards the JSON protocol to the Thoth-OP Server, another specialized SCP service node dedicated to execution planning. The Thoth-OP Server decomposes the structured protocol into atomic operations and generates device-specific command sequences. Each command (for example, a pipetting instruction) is output in a standardized format. The SCP system then dispatches these commands to the appropriate laboratory instruments via their APIs. Throughout this process, the user sees the entire workflow seamlessly managed by SCP: the initial request is transformed into a protocol and executed automatically by the Thoth nodes. 

\begin{figure}[th]
    \centering 
    \includegraphics[width=\textwidth]{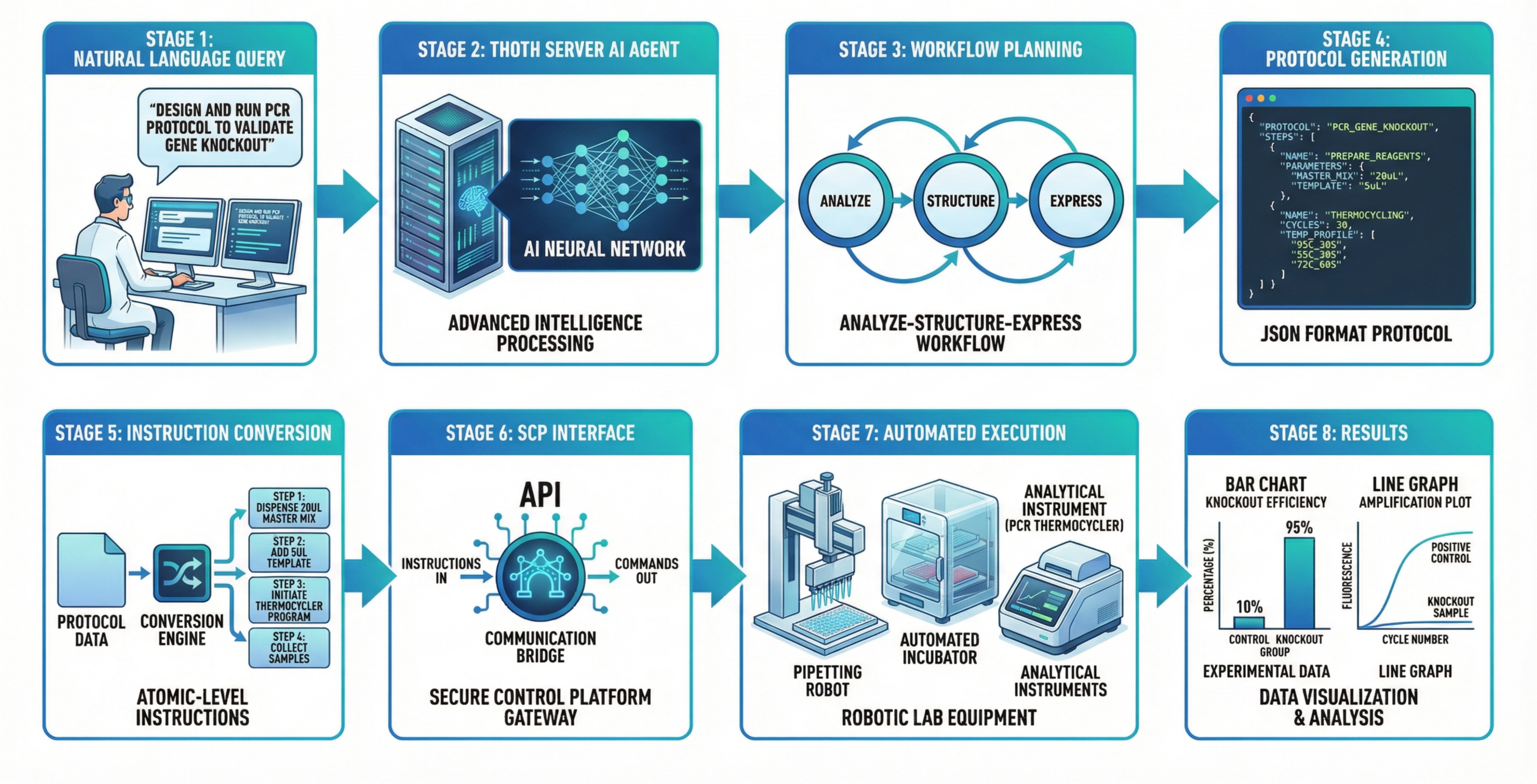} 
    \caption{Case Study 1: Automated Experimental Protocol Design and Execution.} 
    \label{fig:case1} 
\end{figure}

\subsection{Case Study 2: Automated Reproduction of an Existing Protocol from PDF}

In this scenario, SCP is used to automatically reproduce an experiment from an existing protocol document. A researcher does not need to rewrite or adapt the procedure manually: they simply upload a protocol PDF (for example, a method section from a paper or a lab SOP exported from an ELN) to the SCP system. The SCP hub classifies the task as protocol ingestion and routes the document to the Thoth Server, which functions as an SCP-compatible protocol understanding node.

Thoth parses the free-form PDF, extracting the experimental objective, materials, and each step’s actions, parameters, and dependencies. Even when the document uses narrative language or heterogeneous formatting, Thoth converts it into a standardized JSON protocol object that precisely enumerates the workflow: which reagents are used, in what volumes, at which temperatures, for how long, and in what order. From the user’s perspective, “any reasonable wet-lab protocol PDF” is thus transformed into a machine-readable, executable representation without manual transcription.

Once the structured protocol has been generated and optionally reviewed, the SCP hub forwards it to the Thoth-OP execution node for operation planning. Thoth-OP decomposes each step into atomic device-level instructions and validates all parameters against the capabilities and safety limits of the available instruments. It then emits a sequence of standardized commands that the SCP infrastructure dispatches to the appropriate lab devices. As a result, the experiment described in the original PDF is automatically reproduced on the automation platform: pipetting, mixing, incubation, and measurement steps are executed end-to-end under SCP’s control. The researcher experiences the pipeline as: upload a protocol file, confirm the interpreted plan, and watch the system intelligently and safely replay the experiment on real hardware, demonstrating SCP’s ability to turn arbitrary protocol documents into fully automated, reproducible workflows.

\begin{figure}[th]
    \centering 
    \includegraphics[width=\textwidth]{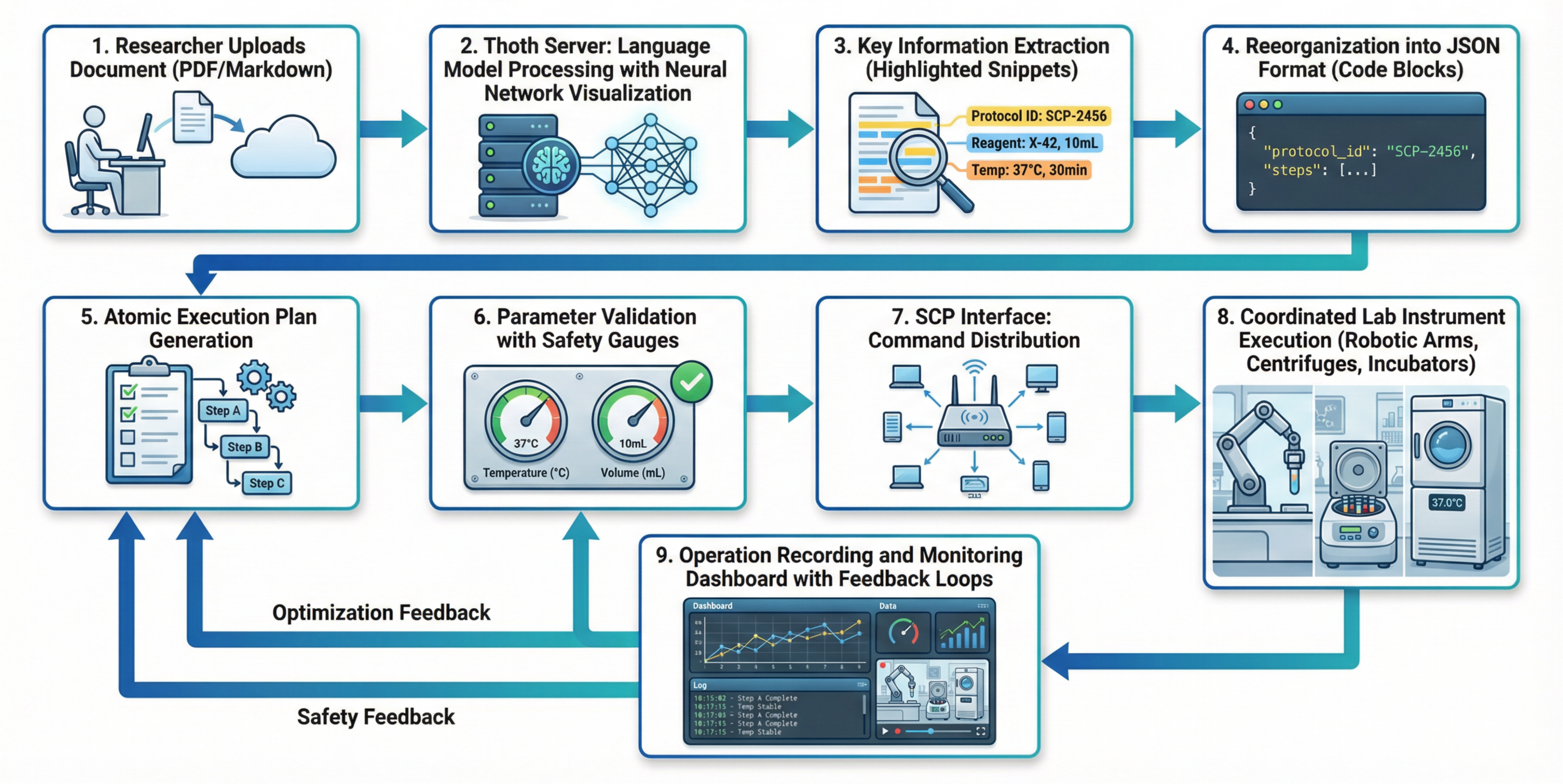} 
    \caption{Case Study 2: Automated Reproduction of an Existing Protocol from PDF.} 
    \label{fig:case2} 
\end{figure}

\subsection{Case Study 3: AI-Driven Molecular Screening and Docking via SCP}

This case study demonstrates how the SCP toolchain can support an end-to-end small-molecule
screening and protein–ligand docking campaign in an automated manner. The workflow
combines cheminformatics, ADMET prediction, structural biology, and molecular docking,
all orchestrated through a unified SCP pipeline.

\paragraph{Step 1: Molecule property evaluation.}
The process starts from a library of 50 small molecules encoded in SMILES format. Using
SCP tools such as \texttt{calculate\_mol\_drug\_chemistry}, the system computes the
QED (Quantitative Estimate of Drug-likeness) score for each molecule, providing a numerical
measure of drug-likeness. In parallel, the \texttt{pred\_molecule\_admet} tool predicts the
LD\textsubscript{50} toxicity metric for each compound. These two quantities jointly
characterize how promising and safe each candidate is from a medicinal chemistry
perspective.

\paragraph{Step 2: Initial filtering of candidate molecules.}
Based on pre-defined thresholds, the workflow filters molecules with QED~$\ge 0.6$ and
LD\textsubscript{50}~$\ge 3.0$. This filtering step eliminates compounds with poor
drug-likeness or high predicted toxicity and yields a refined set of six candidate molecules.
From a user’s perspective, this entire stage is expressed as a small number of SCP tool
invocations over the input SMILES list, and the platform automatically returns a structured
table of qualified compounds.

\paragraph{Step 3: Protein target preparation.}
Next, the workflow prepares the receptor protein for docking. Using PDB ID~6vkv as the
target, the SCP toolchain performs a series of automated operations:
(i) download the PDB structure;
(ii) extract the main chain (default chain~A);
(iii) repair missing or inconsistent regions using PDBFixer; and
(iv) identify putative binding pockets with \texttt{fpocket}.
The pocket with the highest score is selected as the docking site, and its coordinates are
recorded for subsequent stages. All of these operations are invoked as SCP tools, so the
user does not manually handle PDB manipulation or pocket detection; instead, they see a
clean, standardized receptor object with an associated binding region.

\paragraph{Step 4: Format conversion for docking.}
To interface with an AutoDock Vina-style docking engine, both ligands and receptor must be
converted into PDBQT format. The six filtered SMILES molecules are transformed into 3D
structures and exported as ligand PDBQT files, while the repaired protein structure is
likewise converted into a receptor PDBQT file. This conversion is triggered by SCP tools
that hide the underlying cheminformatics steps, ensuring that all input files for docking are
generated in a consistent, reproducible manner.

\paragraph{Step 5: Docking and final hit selection.}
The final stage invokes the \texttt{quick\_molecule\_docking} tool at the previously identified
pocket center. Each of the six candidate ligands is docked into the binding site, and the
workflow records the predicted binding affinities. Molecules with docking affinity
$\le -7.0$~kcal/mol are retained as high-potential hits. In this example, two compounds
survive this final filter:
\begin{itemize}
  \item \texttt{O=C(c1ccc(F)cc1F)N1CCN2C(=O)c3ccccc3C12c1ccc(Cl)cc1}
  \item \texttt{O=C(c1cccc(F)c1)N1CCN2C(=O)c3ccccc3C12c1ccc(Cl)cc1}
\end{itemize}
From the user’s perspective, the SCP pipeline transforms an initial list of 50 SMILES strings
into a pair of prioritized hit molecules bound to a structurally prepared protein target. Each
stage of the workflow—property prediction, filtering, protein preparation, format conversion,
and docking—is encapsulated as composable SCP tools, illustrating how the platform
supports an AI-driven early drug discovery paradigm with minimal manual intervention.

\begin{figure}[th]
    \centering 
    \includegraphics[width=\textwidth]{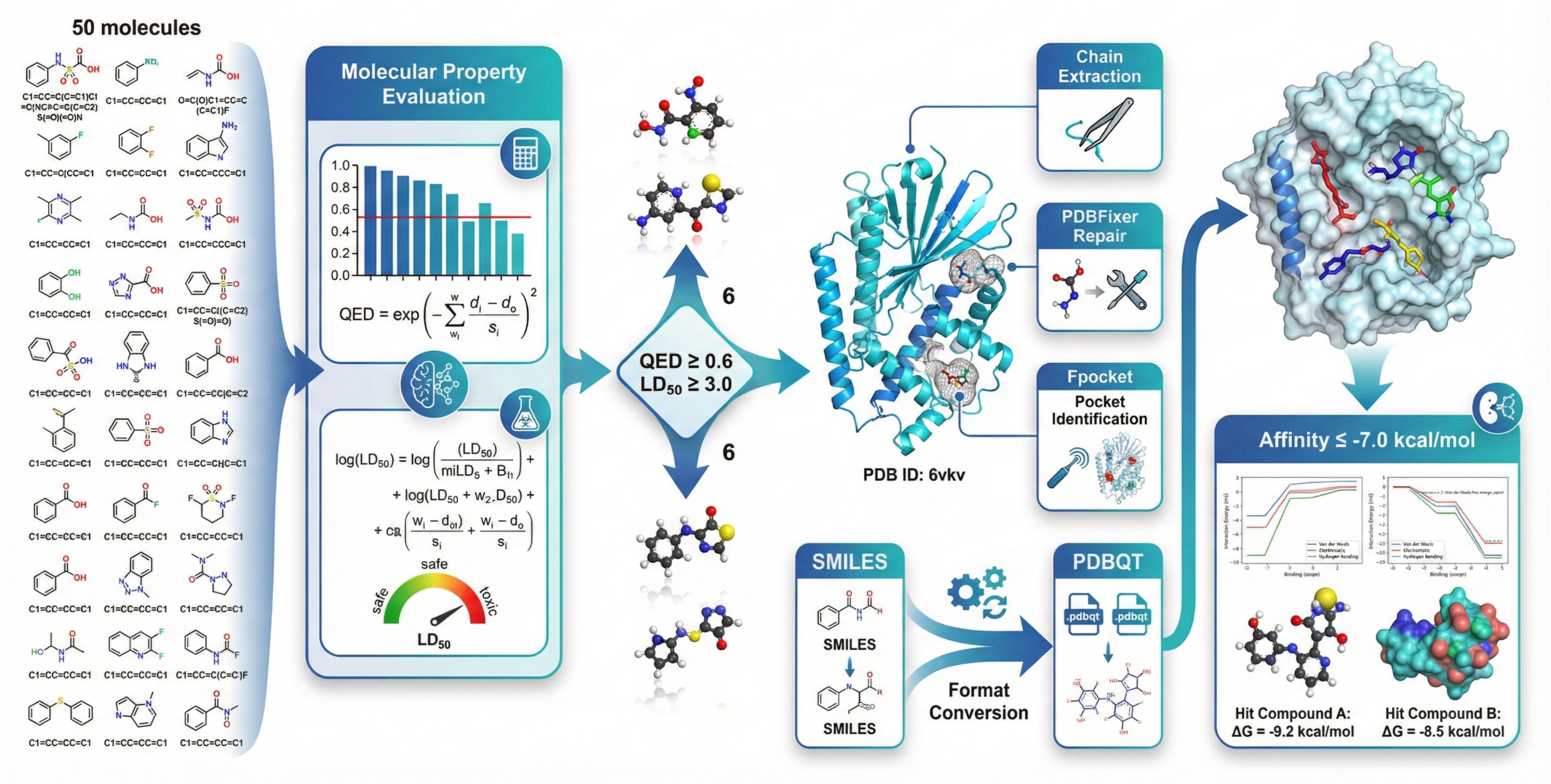} 
    \caption{Case Study 3: AI-Driven Molecular Screening and Docking via SCP.} 
    \label{fig:case3} 
\end{figure}

\subsection{Case Study 4: AI-Assisted Fluorescent Protein Engineering with Dry–Wet Integration via SCP}

This case study illustrates how SCP supports a tightly coupled dry–wet workflow for engineering fluorescent proteins. 

On the ``dry'' side, the scientist begins by specifying an optimization goal in the SCP client, such as increasing brightness or photostability of a given fluorescent protein scaffold under specific experimental conditions. The request is sent to the SCP Hub, which orchestrates a set of computational servers (e.g., sequence design, structural modeling, and property prediction tools) through standardized JSON task plans. These tools explore sequence space around the wild-type protein, perform in silico mutational scanning, and predict key properties such as folding stability, spectral shift, and expression level. The SCP Hub aggregates these predictions into a ranked list of candidate variants and encodes each design, together with its predicted properties and intended assay conditions, into a unified SCP experiment plan that can be directly reused by downstream wet-lab components.

On the ``wet'' side, the same SCP plan is automatically translated into executable experimental workflows. For each selected fluorescent protein variant, the SCP Hub dispatches structured JSON sub-plans to wet-lab servers, which generate detailed protocols for plasmid construction, transformation, cell culture, induction, and fluorescence readout. These protocols are further compiled into atomic operation sequences (such as pipetting, incubation, centrifugation, and plate reading) that can be executed on robotic platforms. During execution, the SCP Hub monitors instrument status and assay progress, streaming quantitative fluorescence measurements and quality-control metrics back into the shared SCP context. This creates a closed loop in which in silico design results, wet-lab measurements, and protocol variants are all represented in a single, standardized SCP timeline. As a result, the fluorescent protein engineering workflow depicted in the figure is no longer a loose combination of simulation and experiment, but a unified dry–wet pipeline: SCP provides the orchestration layer that connects AI-based design, automated experimentation, and iterative optimization into a single, reproducible scientific process.

\begin{figure}[th]
    \centering 
    \includegraphics[width=\textwidth]{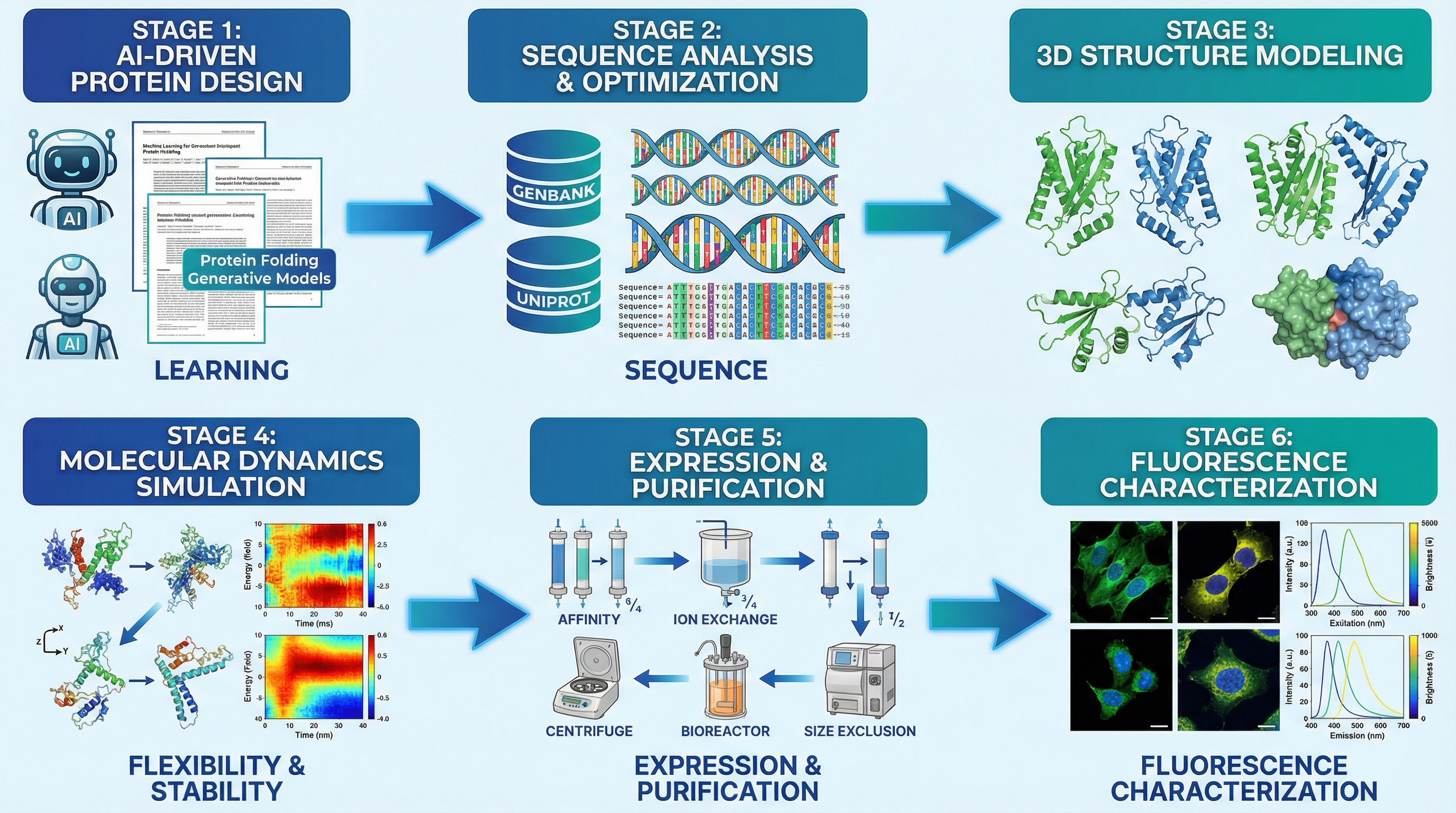} 
    \caption{Case Study 4: AI-Assisted Fluorescent Protein Engineering with Dry–Wet Integration via SCP} 
    \label{fig:case4} 
\end{figure}

\section{Discussion: Comparative Analysis of SCP and MCP}
SCP and MCP~\cite{anthropicModelContextProtocol2024} represent two distinct approaches to orchestrating AI-driven research workflows. MCP provides a general-purpose standard for connecting AI models with tools and data sources. This generic design has been widely adopted for integrating LLMs with software APIs, databases, and other resources in a uniform way. However, when applying MCP to complex scientific domains, several limitations become evident. In contrast, SCP is a specialized framework built to address these shortcomings by introducing domain-specific structure and coordination mechanisms on top of the MCP paradigm \cite{opensciencelabSCP2025}. In this section, we provide a rigorous comparison of MCP and SCP, focusing on MCP's limitations in protocol standardization, high-throughput experimentation, multi-agent orchestration, and integration of wet-lab equipment, and how SCP's design offers solutions in each of these areas.

\paragraph{Protocol Standardization and Contextual Structure.} MCP was designed to standardize basic interactions (e.g., file I/O, function calls) between AI agents and tools across disparate systems. This works well for straightforward tasks (such as querying a database or calling a cloud API)~\cite{replit2025,microsoftCopilotStudioMCP2025,jetbrainsMCPServer2025,theiaAITheiaIDE2025,cloudflare2025,blockSquareMCPTeamTools2025}, but MCP by itself does not impose a structured format for complete scientific protocols or experimental plans. In scientific workflows, researchers require a high-level representation of the entire experiment (including objectives, parameters, and expected outcomes) that can be understood and shared among multiple agents and tools. MCP lacks this notion of a scientific context. It treats interactions as isolated messages rather than parts of a cohesive experiment plan. As a result, protocol standardization in a scientific sense (i.e., a consistent way to describe experiments end-to-end) is weak under MCP: different labs or projects might develop their own conventions on top of MCP, leading to fragmentation.

SCP directly tackles this issue by providing a standardized research workflow representation built into the protocol. Borrowing inspiration from MCP's general interface, SCP defines a structured, JSON-based schema for planning and describing experiments. Each experiment in SCP carries rich metadata (e.g., unique experiment IDs, researcher identifiers, experiment type as dry/wet/hybrid, objectives and so on) and uses a formal JSON schema to outline the procedure steps and resources involved. This structured protocol planning ensures that every step of an experiment is explicit and machine-interpretable. For example, an experiment plan in SCP might specify the sequence of actions (data preprocessing, model training, hypothesis generation, wet-lab validation steps, etc.) in a nested JSON format, which the SCP system can parse and manage. The presence of a well-defined schema means that all agents and tools in the SCP ecosystem interpret the protocol in the same way, greatly enhancing standardization over what MCP alone can achieve. In practice, this means a laboratory adopting SCP can have all their instruments and AI agents follow the same protocol blueprint for any given experiment, reducing ambiguity and setup overhead. In summary, while MCP provides the low-level message format for tool access, SCP adds a higher-level grammar for scientific experimentation, bringing protocol uniformity to complex workflows.

\paragraph{High-Throughput Experimentation Support.} Scientific research often demands running many experiments or iterative trials in order to explore large parameter spaces (for instance, screening dozens of candidate compounds or running thousands of simulations in a materials study). In such high-throughput scenarios, MCP's lack of built-in experiment management becomes a bottleneck. Under MCP, each tool invocation is essentially stateless and context-agnostic, so coordinating a batch of 100 experiments requires external scheduling logic and careful tracking by the user or a separate program. There is no native concept of an “experiment queue” or a systematic way to chain results from one run into the next; the protocol does not remember past actions unless the agent itself implements memory. This limitation means researchers must implement additional layers on top of MCP to handle batching, concurrency, and result aggregation. The absence of a unified context across runs can lead to repeated setup overhead and potential errors when scaling up to high-throughput workflows.
SCP offers enhancements geared towards high-throughput and iterative experimentation. Because SCP treats the entire workflow as a first-class object (with an experiment ID and a persistent context), multiple runs or trial variants can be managed under a common protocol umbrella. Concretely, SCP can automatically log the outcome of each high-throughput experiment instance (e.g., each combination of parameters in a grid search or each candidate molecule in a screening assay) with a uniform format, making it easy to compare and aggregate results. The structured metadata (including priority levels and unique identifiers for each run) allows the system to schedule experiments efficiently and in parallel where possible. Researchers can request an entire batch of experiments in one protocol submission (thanks to the JSON-based plan encoding all trials), and the SCP framework will coordinate their execution and collect all results. SCP improves support for high-throughput and large-scale experimentation by introducing context-aware batching and automated workflow execution, reducing the manual burden present in an MCP-only approach.

\paragraph{Multi-Agent Orchestration and Coordination.} Modern scientific projects increasingly involve multi-agent systems—multiple AI agents (and humans) collaborating, each with specialized roles (e.g., a data-analysis agent, a hypothesis-generating agent, a robotic lab assistant, etc.). MCP, by design, is message-centric but not inherently multi-agent: it defines how an AI agent (client) communicates with external tools (servers), but it does not specify how multiple autonomous agents should coordinate with each other in a complex task. In fact, auxiliary protocols like Google's A2A (Agent-to-Agent) were needed to handle direct inter-agent communication in the absence of a higher-level framework \cite{Google2025}. Without an overarching orchestration mechanism, an MCP-based multi-agent system can devolve into a collection of point-to-point message exchanges with no global oversight. This makes it challenging to implement structured teamwork or division of labor among agents using MCP alone. In essence, MCP offers the messaging pipes, but the orchestration logic for multi-agent workflows in scientific settings is absent. This was recognized as a shortcoming in emerging AI Scientist ecosystems: existing protocols lacked a unified approach to coordinate diverse agents and tools within a single, reasoning-aware workflow. This leads to situations where human scientists must act as the “protocol” between AI agents, manually integrating their outputs.

The SCP framework incorporates multi-agent coordination as a core feature rather than an afterthought. It introduces a centralized SCP Hub that serves as an orchestrator for all participating agents and resources in an experiment. Through the Hub, agents do not just send blind messages; instead, each agent’s actions are contextualized within a shared experiment state managed by the hub. This extension allows the system to decompose high-level goals into sub-tasks, assign those sub-tasks to the appropriate specialized agents, and sequence their execution. The hub also tracks dependencies and completion status of tasks, enabling conditional branching or iteration in protocols (capabilities that raw MCP lacks). Overall, SCP transforms what would be a set of independent MCP conversations into a coordinated conversation among many agents within a single experimental narrative.

\paragraph{Integration of Wet-Lab Equipment and Complex Workflows.} The most distinguishing requirement of scientific protocols (versus generic software tasks) is the integration of wet-lab experiments — interactions with physical laboratory instruments, robots, or assays. MCP was not specifically developed with lab hardware in mind; it can technically wrap instrument commands as tool APIs, but MCP does not define any standards for how lab equipment should be represented or controlled. This lack of built-in support makes integrating wet-lab steps cumbersome: every new piece of hardware might need a custom MCP adapter, and there is no universal format for lab actions (contrast this with how MCP standardized file access or HTTP requests). Moreover, scientific workflows are complex hybrids of dry (computational) and wet (experimental) steps. Under an MCP-only regime, an AI agent might plan an experiment and then output a text protocol for a human to carry out in the lab, because the agent cannot directly interface with lab gear in a standardized way. This gap significantly slows the cycle of experimentation.

SCP addresses wet-lab integration as a first-class objective. In addition to its experiment planning schema, SCP introduces explicit provisions for controlling laboratory devices and incorporating their outputs into the workflow. Specifically, SCP defines standardized device drivers and vendor-agnostic interfaces for common classes of lab equipment, much like instrument drivers in an operating system but adapted to the MCP-style JSON protocol. By having a uniform interface, an SCP-compliant lab instrument (e.g., a thermocycler, HPLC machine, or robotic arm) can be invoked in an experiment plan just as easily as a computational tool. The SCP Hub and associated edge servers handle the low-level communication with devices, so an AI agent can simply call a high-level action and the SCP middleware will translate that into the specific device commands required, irrespective of manufacturer differences. This dramatically lowers the barrier to including wet-lab steps in an automated workflow.

\section{Conclusion}
SCP (Scientific Context Protocol) provides an open, standardized \emph{protocol layer} for connecting and orchestrating heterogeneous scientific resources---including \textbf{1,600+} software tools, models (including LLMs), datasets, workflow engines, and wet-lab instruments---under a unified, interoperable interface. By turning fragmented research components into composable \emph{SCP Servers} that can be reliably discovered, invoked, and combined, SCP reduces integration overhead and enables reproducible, traceable end-to-end workflows spanning dry and wet laboratories. Moreover, a centralized \emph{SCP Hub} maintains persistent scientific context, enforces per-experiment authentication and authorization, and manages the experiment lifecycle (planning, execution, monitoring, and archival), enabling secure cross-institution collaboration and scalable orchestration for autonomous AI scientists. Overall, SCP transforms isolated agents, tools, and instruments into an interoperable ``web'' of capabilities that can be safely composed as discovery services, laying foundational infrastructure for a more connected, extensible, and accelerated paradigm of agentic scientific discovery.

\bibliography{ref}
\bibliographystyle{plain}

\appendix
\section{Complete SCP Server List}
\makeatletter
\@ifpackageloaded{longtable}{}{\usepackage{longtable}}
\@ifpackageloaded{array}{}{\usepackage{array}}
\@ifpackageloaded{booktabs}{}{\usepackage{booktabs}}
\@ifpackageloaded{xcolor}{}{\usepackage{xcolor}}
\@ifpackageloaded{hyperref}{}{\usepackage{hyperref}}
\makeatother

\newcolumntype{L}[1]{>{\raggedright\arraybackslash}p{#1}}

\begin{center}
\footnotesize
\renewcommand{\arraystretch}{1.2}
\begin{longtable}{
  L{2.6cm}   
  L{1.4cm}   
  L{3.6cm}   
  L{3.0cm}   
  c        
}
\label{tab:scpserver}\\
\caption{Summary of current SCP servers}\\\toprule
\textbf{Server name} & \textbf{Domain} & \textbf{Description} & \textbf{Endpoint} & \textbf{Tool count}\\ \midrule
\endfirsthead

\midrule
\endhead

\midrule
\multicolumn{5}{r}{\footnotesize}\\
\endfoot
\bottomrule
\endlastfoot

VenusFactory & Biology &
VenusFactory is an AI-centric protein-engineering platform that unifies code, notebook, GUI and agent workflows around Venus and ESM protein models, covering mutation and function prediction, residue screening, data retrieval, model training, evaluation and deployment. &
\url{https://scp.intern-ai.org.cn/api/v1/mcp/1/VenusFactory} & 12\\
\hline
DrugSDATool & Biology &
DrugSDATool is an integrated toolkit for drug screening, design and analysis that bundles Open Babel, RDKit and BioPython functions to retrieve data, interconvert formats, repair protein structures, normalize ligands, parse molecules, compute similarity and analyze binding pockets. &
\url{https://scp.intern-ai.org.cn/api/v1/mcp/2/DrugSDATool} & 33\\
\hline
DrugSDAModel & Biology &
DrugSDAModel unites docking, pocket detection, affinity and ADMET prediction, protein structure modeling and disease-reversal scoring into one suite for end-to-end AI-driven drug discovery. &
\url{https://scp.intern-ai.org.cn/api/v1/mcp/3/DrugSDAModel} & 8\\
\hline
OrigeneChEMBL & Biology, Chemistry &
Origene embeds the full ChEMBL engine for instant, on-demand querying of small-molecule bioactivity data. &
\url{https://scp.intern-ai.org.cn/api/v1/mcp/4/OrigeneChEMBL} & 58\\
\hline
OrigeneKEGG & Biology &
Origene embeds the full KEGG engine for instant, on-demand pathway and functional annotation queries. &
\url{https://scp.intern-ai.org.cn/api/v1/mcp/5/OrigeneKEGG} & 6\\
\hline
OrigeneSTRING & Biology &
Origene embeds the full STRING engine for instant, on-demand protein interaction and functional annotation queries. &
\url{https://scp.intern-ai.org.cn/api/v1/mcp/6/OrigeneSTRING} & 8\\
\hline
OrigeneSearch & Biology &
Origene seamlessly integrates Tavily, Jina, ClinVar, GSEA, PubMed and other leading sources for one-click search and instant access. &
\url{https://scp.intern-ai.org.cn/api/v1/mcp/7/OrigeneSearch} & 7\\
\hline
OrigenePubChem & Chemistry &
Origene embeds the full PubChem engine for instant, on-demand access to the world’s largest open chemical database. &
\url{https://scp.intern-ai.org.cn/api/v1/mcp/8/OrigenePubChem} & 39\\
\hline
OrigeneNCBI & Biology &
Origene embeds the full NCBI engine for instant, on-demand access to the world’s largest open biological database. &
\url{https://scp.intern-ai.org.cn/api/v1/mcp/9/OrigeneNCBI} & 52\\
\hline
OrigeneUniProt & Biology &
Origene embeds the full UniProt engine for instant, on-demand access to the authoritative protein sequence and function database. &
\url{https://scp.intern-ai.org.cn/api/v1/mcp/10/OrigeneUniProt} & 23\\
\hline
OrigeneTCGA & Biology &
Origene embeds the full TCGA engine for instant, on-demand access to the authoritative cancer genomics atlas. &
\url{https://scp.intern-ai.org.cn/api/v1/mcp/11/OrigeneTCGA} & 3\\
\hline
OrigeneEnsembl & Biology &
Origene embeds the full Ensembl engine for instant, on-demand access to the authoritative genome annotation and comparative database. &
\url{https://scp.intern-ai.org.cn/api/v1/mcp/12/OrigeneEnsembl} & 96\\
\hline
OrigeneUCSC & Biology &
Origene embeds the full UCSC Genome Browser engine for instant, on-demand genome annotation and visualization. &
\url{https://scp.intern-ai.org.cn/api/v1/mcp/13/OrigeneUCSC} & 9\\
\hline
OrigeneFDADrug & Biology &
Origene embeds the full FDA Drug engine for instant, on-demand access to authoritative drug regulatory and approval data. &
\url{https://scp.intern-ai.org.cn/api/v1/mcp/14/OrigeneFDADrug} & 155\\
\hline
OrigeneOpenTargets & Biology &
Origene embeds the full Open Targets engine for instant, on-demand target discovery and validation. &
\url{https://scp.intern-ai.org.cn/api/v1/mcp/15/OrigeneOpenTargets} & 92\\
\hline
OrigeneMonarch & Biology &
Origene embeds the full Monarch Initiative engine for instant, on-demand disease-phenotype-gene associations. &
\url{https://scp.intern-ai.org.cn/api/v1/mcp/16/OrigeneMonarch} & 3\\
\hline
BioInfoTools & Biology &
BioInfoTools is a plug-and-play protein sequence analysis service that wraps InterProScan and BLAST into one unified API for domain detection, GO annotation and similarity search. &
\url{https://scp.intern-ai.org.cn/api/v1/mcp/17/BioInfoTools} & 3\\
\hline
ThothOP & Biology &
ThothOP is the wet-lab execution engine that exposes stable atomic commands for pipetting, mixing, incubation, centrifugation and more, letting users or agents compose and run protocols safely and precisely. &
\url{https://scp.intern-ai.org.cn/api/v1/mcp/18/ThothOP} & 58\\
\hline
ThothPlan & Biology &
ThothPlan is a reinforcement learning fine tuned LLM agent that turns research ideas into executable wet lab protocols, eliminating manual scripting and delivering accurate, logically sound and fully automated experiments. &
\url{https://scp.intern-ai.org.cn/api/v1/mcp/19/ThothPlan} & 4\\
\hline
Materials Mechanics and Fracture Analysis & Mechanics and Materials Science &
Materials \& Fracture Analysis-Tool is a one-stop library that unites stress-strain, fracture criteria, safety factors, elastic-plastic parameters, interface strength and residual stress calculations for rapid, accurate structural and failure analysis. &
\url{https://scp.intern-ai.org.cn/api/v1/mcp/20/Materials\_Mechanics\_and\_Fracture\_Analysis} & 107\\
\hline
Electrical Engineering and Circuit Calculations & Physics &
Electrical \& Circuit Calculation-Tool is a unified utility library that spans basic electrical quantities to advanced circuit simulation, covering DC/AC analysis, series-parallel equivalents, critical current density, magnetic field strength, electromagnetic quantification, error validation and duty-cycle detection for rapid design and experiment support. &
\url{https://scp.intern-ai.org.cn/api/v1/mcp/21/Electrical\_Engineering\_and\_Circuit\_Calculations} & 73\\
\hline
Thermal Fluid Dynamics & Physics &
ThermoFluid-Tool is an integrated library for engineering thermodynamics and fluid mechanics, bundling energy conservation, heat transfer, phase change, pressure, velocity and property estimation to enable rapid numerical analysis, modeling and data processing of complex thermal-fluid systems. &
\url{https://scp.intern-ai.org.cn/api/v1/mcp/22/Thermal\_Fluid\_Dynamics} & 77\\
\hline
Optics and Electromagnetics & Physics &
Optics \& Electromagnetics-Tool is a general-purpose library that unifies light propagation, refraction, reflection, interference, electromagnetic wave parameters and field-matter interactions for lens design, optical simulation, field strength, energy density and experimental data processing. &
\url{https://scp.intern-ai.org.cn/api/v1/mcp/23/Optics\_and\_Electromagnetics} & 30\\
\hline
Chemistry and Reaction Calculations & Chemistry &
Chemistry \& Reaction Calculation-Tool is a unified toolkit that integrates reaction stoichiometry, concentration, rate and equilibrium constants, mass and energy balancing, yield estimation and solution preparation for rapid experimental and process design support. &
\url{https://scp.intern-ai.org.cn/api/v1/mcp/24/Chemistry\_and\_Reaction\_Calculations} & 105\\
\hline
Geometry and mathematical calculations & Mathematics &
Geometry \& Math Calculation-Tool is a general-purpose library bundling numerical operations, algebraic derivation, function fitting, statistical analysis and spatial geometry for universal modeling, computing and engineering design. &
\url{https://scp.intern-ai.org.cn/api/v1/mcp/25/Geometry\_and\_mathematical\_calculations} & 99\\
\hline
Data processing and statistical analysis & Information Science and Computing Technology &
Data Processing \& Statistical Analysis-Tool is a hands-on library that unifies cleaning, filtering, normalization, outlier detection, fitting, interpolation, error assessment, distribution and correlation analysis for research, engineering and data-driven decisions. &
\url{https://scp.intern-ai.org.cn/api/v1/mcp/26/Data\_processing\_and\_statistical\_analysis} & 41\\
\hline
Physical Quantities Conversion & Information Science and Computing Technology &
Physical Quantities \& Unit Conversion-Tool is an all-in-one library for converting common and specialized units, swapping SI and imperial systems, fetching physical constants, adjusting magnitudes and validating dimensions for seamless scientific and engineering calculations. &
\url{https://scp.intern-ai.org.cn/api/v1/mcp/27/Physical\_Quantities\_Conversion} & 81\\
\hline
InternAgent & Information Science and Computing Technology &
InternAgent builds a suite of over 100 cross-domain scientific tools and deploys InternAgent-DeepResearch, an RL-tuned agent that decomposes complex tasks into dependent subtasks, explores them in parallel and self-optimizes to produce rigorous, fully referenced research reports. &
\url{https://scp.intern-ai.org.cn/api/v1/mcp/28/InternAgent} & 110\\
\hline
SciToolAgent-Bio & Biology &
SciToolAgent-Bio is a one-stop toolkit covering proteomics and genomics: it computes protein properties, optimizes codons, predicts folds, aligns sequences, detects signal peptides, designs peptide libraries, maps cleavage sites, labels antibody CDRs, scores solubility and forecasts drug-target interactions. &
\url{https://scp.intern-ai.org.cn/api/v1/mcp/29/SciToolAgent-Bio} & 57\\
\hline
SciToolAgent-Mat & Mechanics and Materials Science &
SciToolAgent-Mat is a materials-science toolkit that parses MOF lattices and topologies, mines Materials Project for bandgaps, elastic and dielectric data, evaluates battery voltage, capacity and cycling life, and computes thermodynamics, vibration, adsorption and stability in one systematic workflow. &
\url{https://scp.intern-ai.org.cn/api/v1/mcp/30/SciToolAgent-Mat} & 8\\
\hline
SciToolAgent-Chem & Chemistry &
SciToolAgent-Chem is a unified cheminformatics suite that interconverts SMILES/InChI/CAS/SELFIES, computes descriptors and fingerprints, predicts reactions and retrosynthesis, scores similarity, identifies functional groups, flags safety hazards, and clusters or classifies molecules with MLP/AdaBoost/RF for end-to-end drug design support. &
\url{https://scp.intern-ai.org.cn/api/v1/mcp/31/SciToolAgent-Chem} & 169\\
\hline
SCP-Workflow & Information Science and Computing Technology &
SCP Workflow is the intelligent coordination hub that exposes a unified, extensible interface to discover, filter and compose any SCP tool on demand, evolving toward adaptive orchestration of wet-lab protocols. &
\url{https://scp.intern-ai.org.cn/api/v1/mcp/32/SCP-Workflow} & 2\\
\end{longtable}
\end{center}

\section{Example code}
The following code demonstrates how to connect to SCP servers and invoke specialized tools to implement Case Study 3. Execution codes of more SCP examples are available in the code repository \url{https://github.com/InternScience/scp}.
\begin{python}
import asyncio
import json
from mcp.client.streamable_http import streamablehttp_client
from mcp import ClientSession

## Server endpoints of DrugSDA-Model and DrugSDA-Tool
DrugSDA_Model_SERVER_URL = "https://scp.intern-ai.org.cn/api/v1/mcp/3/DrugSDA-Model"    
DrugSDA_Tool_SERVER_URL = "https://scp.intern-ai.org.cn/api/v1/mcp/2/DrugSDA-Tool"      

## Definition of the DrugSDA SCP client, including basic operations such as connect, disconnect, list_tools, and parse_result.
class DrugSDAClient:    
    def __init__(self, server_url: str):
        self.server_url = server_url
        self.session = None
        
    async def connect(self):
        print(f"server url: {self.server_url}")
        try:
            self.transport = streamablehttp_client(
                url=self.server_url,
                headers={"SCP-HUB-API-KEY": "sk-xxx"}
            )
            self.read, self.write, self.get_session_id = await self.transport.__aenter__()
            
            self.session_ctx = ClientSession(self.read, self.write)
            self.session = await self.session_ctx.__aenter__()

            await self.session.initialize()
            session_id = self.get_session_id()
            
            print(f"connect success")
            return True
            
        except Exception as e:
            print(f"connect failure: {e}")
            import traceback
            traceback.print_exc()
            return False
    
    async def disconnect(self):
        try:
            if self.session:
                await self.session_ctx.__aexit__(None, None, None)
            if hasattr(self, 'transport'):
                await self.transport.__aexit__(None, None, None)
            print("already disconnect")
        except Exception as e:
            print(f"disconnect error: {e}")
    
    async def list_tools(self):        
        try:
            tools_list = await self.session.list_tools()
            print(f"tool count: {len(tools_list.tools)}")
            
            for i, tool in enumerate(tools_list.tools, 1):
                print(f"{i:2d}. {tool.name}")
                if tool.description:
                    desc_line = tool.description.split('\n')[0]
                    print(f"    {desc_line}")
            
            print(f"Get tool list success")
            return tools_list.tools
            
        except Exception as e:
            print(f"Get tool list fail: {e}")
            return []
    
    def parse_result(self, result):
        try:
            if hasattr(result, 'content') and result.content:
                content = result.content[0]
                if hasattr(content, 'text'):
                    return json.loads(content.text)
            return str(result)
        except Exception as e:
            return {"error": f"parse error: {e}", "raw": str(result)}

## Excution code of calling tools to perform Case Study 3
async def main():
    tool_client = DrugSDAClient(DrugSDA_Tool_SERVER_URL)
    if not await tool_client.connect():
        print("connection failed")
        return
    
    model_client = DrugSDAClient(DrugSDA_Model_SERVER_URL)
    if not await model_client.connect():
        print("connection failed")
        return

    ## Input smiles, we don't show all the testing smiles here 
    smiles_list = [
    'O=C(Nc1cccc2c1CCCC2)N1CCc2c([nH]c3ccccc23)C1c1cccc(F)c1F',
    'Cc1ccccc1N1CCN(C2=Nc3cc(Cl)ccc3Nc3ccc(F)cc32)CC1',
    'O=C(c1ccccc1F)N1CCN2C(=O)c3ccccc3C12c1ccc(Cl)cc1',
    ... ... ... ...
    'O=C(NCc1cccc(-c2cccc(-c3cc4c[nH]ccc-4n3)c2O)c1)Nc1ccc(F)cc1'
    ]

    ## step 1. calculate QED scores, call SCP tool calculate_mol_drug_chemistry
    result = await tool_client.session.call_tool(
        "calculate_mol_drug_chemistry",
        arguments={
            "smiles_list": smiles_list
        }
    )
    result_data = tool_client.parse_result(result)
    QED_result = result_data["metrics"]
    print ("Compute QED score finish ...")
    
    ## step 2. Calculate LD50 scores, call SCP tool pred_molecule_admet
    result = await model_client.session.call_tool(
        "pred_molecule_admet",
        arguments={
            "smiles_list": smiles_list
        }
    )
    result_data = model_client.parse_result(result)
    LD50_result = result_data["admet_preds"]
    print ("Predict admet finish ...")
    
    ## step 3. Filter molecules with QED larger than 0.6 and LD50_Zhu larger than 3.0
    select_smiles_list = []
    for i in range(len(smiles_list)):
        smiles = smiles_list[i]
        QED = QED_result[i]["qed"]
        LD50 = LD50_result[i]["LD50_Zhu"]
        if QED >= 0.6 and LD50 >= 3.0:
            select_smiles_list.append(smiles)
            
    print (len(select_smiles_list), select_smiles_list[0])
    
    ## step 4. Retrieve and download the target protein structure, call SCP tool retrieve_protein_data_by_pdbcode.
    pdb_code = "6vkv"
    result = await tool_client.session.call_tool(
        "retrieve_protein_data_by_pdbcode",
        arguments={
            "pdb_code": pdb_code
        }
    )

    result_data = tool_client.parse_result(result)
    pdb_path = result_data["pdb_path"]
    print ("download protein structure: ", pdb_path)
    
    ## step 5. Extract main chain, call SCP tool save_main_chain_pdb
    result = await tool_client.session.call_tool(
        "save_main_chain_pdb",
        arguments={
            "pdb_file_path": pdb_path,
            "main_chain_id": ""
        }
    )
    
    result_data = tool_client.parse_result(result)
    pdb_path = result_data["out_file"]
    print ("extract protein chain: ", pdb_path)
    
    ## step 6. Fix PDB, call SCP tool fix_pdb_dock
    result = await tool_client.session.call_tool(
        "fix_pdb_dock",
        arguments={
            "pdb_file_path": pdb_path
        }
    )
    
    result_data = tool_client.parse_result(result)
    pdb_path = result_data["fix_pdb_file_path"]
    print ("fix protein pdb: ", pdb_path)
    
    ## step 7. Identify binding pockets, call SCP tool run_fpocket
    result = await model_client.session.call_tool(
        "run_fpocket",
        arguments={
            "pdb_file_path": pdb_path
        }
    )
    
    result_data = model_client.parse_result(result)
    best_pocket = result_data["pockets"][0]
    print ('pocket info: ', best_pocket)
    
    ## step 8. Convert SMILES to PDBQT format, call SCP tool convert_smiles_to_format
    result = await tool_client.session.call_tool(
        "convert_smiles_to_format",
        arguments={
            "inputs": select_smiles_list,
            "target_format": "pdbqt"
        }
    )
    result_data = tool_client.parse_result(result)
    ligand_paths = [x["output_file"] for x in result_data["convert_results"]]
    print (ligand_paths)
    
    ## step 9. Convert receptor PDB to PDBQT format, call SCP tool convert_pdb_to_pdbqt_dock
    result = await tool_client.session.call_tool(
        "convert_pdb_to_pdbqt_dock",
        arguments={
            "pdb_file_path": pdb_path
        }
    )
    result_data = tool_client.parse_result(result)
    receptor_path = result_data["output_file"]
    print (receptor_path)
    
    ## step 10. Molecular docking, call SCP tool quick_molecule_docking
    result = await model_client.session.call_tool(
        "quick_molecule_docking",
        arguments={
            "receptor_path": receptor_path,
            "ligand_paths": ligand_paths,
            "center_x": best_pocket["center_x"],
            "center_y": best_pocket["center_y"],
            "center_z": best_pocket["center_z"],
            "size_x": best_pocket["size_x"],
            "size_y": best_pocket["size_y"],
            "size_z": best_pocket["size_z"]
        }
    )
    result_data = model_client.parse_result(result)
    print ("docking finish ...")
    
    ## step 11. Select final candidate molecules with binding affinity smaller than -7.0 kcal/mol
    final_smiles_list = []
    index = 0
    for item in result_data["docking_results"]:
        affinity = item['affinity']
        if affinity <= -7.0:
            final_smiles_list.append(select_smiles_list[index])
        index += 1
    
    print (final_smiles_list)
    
    await model_client.disconnect()
    await tool_client.disconnect()
    
if __name__ == '__main__':
    await main()
\end{python}
\section{Complete SCP Tool List}
\begin{longtable}{|p{4.5cm}|p{4cm}|p{3cm}|}
\caption{The current list of all available SCP tools.}
\renewcommand{\arraystretch}{1.15}
\label{tab:toollist} \\
\hline
\textbf{Tool Name} & \textbf{Description} & \textbf{Category} \\
\hline
\endfirsthead
\hline
\textbf{Tool Name} & \textbf{Description} & \textbf{Category} \\
\hline
\endhead
\seqsplit{predict\_protein\_function} & Predict protein function. & Computational Tools \\
\hline
\seqsplit{predict\_functional\_residue} & Predict functional residue. & Computational Tools \\
\hline
\seqsplit{is\_valid\_protein\_sequence} & Check if the input protein sequence string is valid. & Computational Tools \\
\hline
\seqsplit{is\_valid\_smiles} & Check if the input SMILES string is valid & Computational Tools \\
\hline
\seqsplit{convert\_smiles\_to\_format} & Convert a list of SMILES strings or a list of SMI file paths into other molecular formats. & Computational Tools \\
\hline
\seqsplit{convert\_pdb\_to\_pdbqt\_dock} & Convert a protein file from PDB format to PDBQT format for docking preparation. & Computational Tools \\
\hline
\seqsplit{convert\_complex\_cif\_to\_pdb} & Convert a protein-ligand complex file from CIF format to PDB format. & Computational Tools \\
\hline
\seqsplit{visualize\_protein} & Visualize the protein structure and save as an image. & Computational Tools \\
\hline
\end{longtable}

\end{document}